\documentclass[a4paper]{article}
\usepackage{INTERSPEECH2022}
\usepackage{amsmath,amsfonts}
\usepackage{algorithmic}
\usepackage{algorithm}
\usepackage{array}
\usepackage[hidelinks]{hyperref}
\usepackage{xcolor}
\usepackage{multirow}
\usepackage{makecell}
\usepackage{subfig}
\usepackage{graphicx} 
\usepackage{bbm}

\DeclareMathOperator*{\argmin}{\arg\!\min\,}
\usepackage{tikz}
\definecolor{red}{HTML}{FF0000}
\definecolor{blue}{HTML}{0000FF}
\definecolor{gray}{HTML}{808080}

\title{Autoregressive Co-Training for Learning Discrete Speech Representations}
\name{Sung-Lin Yeh, Hao Tang}
\address{School of Informatics, University of Edinburgh}
\email{sunglin.yeh@ed.ac.uk, hao.tang@ed.ac.uk}

\newcommand{\term}[1]{\textbf{#1}}

\begin{document}

\maketitle
\begin{abstract}
While several self-supervised approaches for learning discrete speech representation have been proposed,
it is unclear how these seemingly similar approaches relate to each other.
In this paper, we consider a generative model with discrete latent variables that learns
a discrete representation for speech.
The objective of learning the generative model is formulated as information-theoretic co-training.
Besides the wide generality, the objective can be optimized with several approaches,
subsuming HuBERT-like training and vector quantization for learning discrete representation.
Empirically, we find that the proposed approach learns discrete representation
that is highly correlated with phonetic units, more correlated than HuBERT-like training
and vector quantization.\footnote{We will publicly release the implementation of autoregressive co-training at 
\href{https://github.com/30stomercury/autoregressive-co-training}{https://github.com/30stomercury/autoregressive-co-training}.}

\end{abstract}

\section{Introduction}

Self-supervised learning has been widely successful at learning
discrete representations from speech.
For example, many \cite{ref:chorowski2019unsupervised,ref:chung2020vector, ref:baevski2020wav2vec,ref:zhou2021comparison}
have adopted self-supervised representations for acoustic unit discovery.
Besides, constraining certain representations to be discrete appears to encourage
the learning of phonetic information and improves performance
on automatic speech recognition (ASR) \cite{ref:baevski2020wav2vec,ref:hsu2021hubert,ref:ling2020decoar,ref:chung2021w2v}.
The main technique enabling the learning of discrete representations
is \term{vector quantization} (VQ) \cite{ref:mnih2014neural,ref:van2017neural},
where a vector is mapped to the nearest cluster and is represented by the
identity of the cluster.
A sequence of real-valued vectors, the typical representation produced
by self-supervised models, can be discretized into a sequence of discrete
units using vector quantization.

Despite discretization having zero gradient almost everywhere,
Gumbel softmax \cite{ref:maddison2016concrete,ref:jang2016categorical}
makes it possible to optimize self-supervised
objectives end to end, regardless of whether the objective
is reconstruction-based \cite{ref:chung2020vector,ref:ling2020decoar}
or contrastive-based \cite{ref:baevski2020wav2vec,ref:hsu2021hubert}.
In discrete representation learning of speech,
the training of HuBERT \cite{ref:hsu2021hubert} pushes the idea further, computing
the contrastive objective based on quantized acoustic features.
Contrasting quantized acoustic features is reminiscent to
predicting the cluster identities of frames, and we refer
to this approach as \term{HuBERT-like training}.

The combination of various training objectives and where
the quantization layer is placed creates a variety of seemingly
similar training objectives.
In this paper, we propose a generative model with discrete latent
variables, making the relationship among them explicit.
To train this generative model, we formulate training under
\term{information-theoretic co-training}~\cite{ref:mcallester2018information}, noticing
that there are often two players collaborating to optimize an objective.
This is especially common in contrastive approaches, where
one player produces a representation, while another player
predicts the produced representation.
For example, contrastive predictive coding has a convolutional
neural network producing a representation that are predicted
by a recurrent neural network.
The approach has been inherited in wav2vec 2.0,
where a convolutional neural network produces a discrete representation
that is predicted by a Transformer \cite{ref:baevski2020wav2vec}.
HuBERT pushes this further where one player simply quantizes
the acoustic features, while the other player is still
a Transformer \cite{ref:hsu2021hubert}, predicting the quantized
acoustic features.

Based on this observation, it is natural to consider training
our model with co-training.
In co-training \cite{ref:mcallester2018information}, 
the player that predicts the representation is called,
not surprisingly, a \term{prediction network}, while the other player that
produces the representation is called a \term{confirmation network},
confirming the predicted representation.
The two networks see two different views of the input acoustic features,
such as the past versus the future or the masked versus the unmasked,
so as to avoid degenerate solutions when the two players collaborate.
For example, the prediction network takes the past as input and predicts
a representation of the future, while the confirmation network takes
the future and confirms the predicted representation. 
In this paper, we will follow the past and future views,
i.e., an \term{autoregressive co-training} approach, as it is more
intuitive and consistent with the original work of McAllester \cite{ref:mcallester2018information}.
We will leave the extension of masked and unmasked view to future work.

Our approach is similar to but more general than autoregressive predictive
coding (APC) \cite{ref:chung2019unsupervised}.
VQ-APC \cite{ref:chung2020vector} can be seen as a special case under
our forumulation.
We will also show that HuBERT-like training is equivalent to performing block-coordinate
descent on our co-training objective.
Instead of limiting ourselves to this particular optimization approach,
we will empirically show that performing stochastic gradient descent
directly on the co-training objective greatly improves optimization.

To study the utility of the representation, we evaluate the learned
representation with phone classification, because the task shows
the amount of phonetic information accessible by a linear classifier,
and has been shown to correlate well with applications such as
ASR \cite{ref:maas2017building, ref:chung2020generative}.
We compare several representation learning approaches, APC, VQ-APC,
HuBERT-like training, and co-training.
We find that directly optimizing the co-training objective
significantly improves the phone classification results.

To summarize, we make the following contributions.
First, we propose a generative model with discrete latent variables
for learning discrete speech representations.
Second, we provide several alternatives to train the generative model,
in particular, information-theoretic co-training which subsumes
VQ-APC and HuBERT-like training.
Finally, we show that directly optimizing the co-training objective
significantly improves the optimization of the objective and the
quality of the representation measured by phone classification.

\section{Autoregressive Co-Training}

Consider an utterance of $T$ frames $x_1, x_2, \dots, x_T$.
For $1 \leq t \leq T$, each frame $x_t$ is a real-valued vector, for example, a log Mel spectrum.
For every anchor time point $t$, we refer to frames before $t$ as the past,
and frames after $t$ as the future.
We first assume each future frame $x_{t+k}$ for some $k > 0$ is governed by a latent
variable $z_{t+k}$, meaning that $x_{t+k}$ is independent of everything
else given $z_{t+k}$.
The latent variable $z_{t+k} \in \{1, 2, \dots, N\}$ is discrete and
is said to be the \term{discrete representation} of the frame $x_{t+k}$.
We also assume that $z_{t+k}$ depends on frames from the past $x_{1:t}$.\footnote{
We use $x_{1:t}$ as a shorthand for $x_1, x_2, \dots, x_t$.}

\begin{figure}
\centering\includegraphics{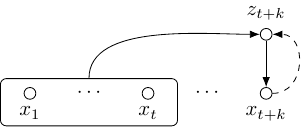}
\caption{
    The proposed graphical model for generating acoustic frames.
    Solid lines denote the generative process,
    while the dashed line denotes the auxiliary distribution $q(z_{t+k}|x_{t+k})$.
}
\label{fig:1}
\end{figure}

The graphical model based on our assumptions is shown in Figure \ref{fig:1}.
Training this graphical model amounts to learning the distribution $p(z_{t+k} | x_{1:t})$
and $p(x_{t+k} | z_{t+k})$.
The likelihood of the parameters can be written as
\begin{align}
    p(x_{1:T}) = p(x_{1:k}) \prod_{t=1}^{T-k} p(x_{t+k} | x_{1:t}). \label{eq:likelihood}
\end{align}
We can further write
\begin{align}
    p(x_{t+k}|x_{1:t}) = \sum_{z_{t+k}} p(x_{t+k}|z_{t+k}) p(z_{t+k}|x_{1:t}),
\label{eq:future}
\end{align}
where the factorization is based on the independence assumptions
in the graphical model.
The model is reminiscent to autoregressive predictive coding (APC) \cite{ref:chung2019unsupervised}.
In fact, we will discuss how VQ-APC \cite{ref:chung2020vector} is a special case of our model.

Given the generative process, we propose to train the model with
information-theoretic co-training \cite{ref:mcallester2018information} instead of the
likelihood \eqref{eq:likelihood}.
Below we briefly review information-theoretic co-training
under our setting.

The goal of co-training is to maximize the mutual information
of the past $x_{1:t}$ and the discrete representation $z_{t+k}$
had we know the future frame $x_{t+k}$.
We introduce an auxiliary distribution $q(z_{t+k} | x_{t+k})$
to simplify inference.
By the definition of mutual information, we have that
\begin{align}
I(z_{t+k}, x_{1:t} | x_{t+k}) = H(z_{t+k} | x_{t+k}) - H(z_{t+k} | x_{1:t}, x_{t+k}). \label{eq:mi}
\end{align}
The key observation is that we can upper bound
entropy with cross entropy and get
\begin{align}
    H(z_{t+k} | x_{1:t}, x_{t+k}) &= \mathbb{E}_{z_{t+k} \sim q}[-\log q(z_{t+k} | x_{1:t}, x_{t+k})] \notag\\
        & \leq \mathbb{E}_{z_{t+k} \sim q}[-\log p(z_{t+k} | x_{1:t}, x_{t+k})], \label{eq:ce}
\end{align}
where we have
\begin{align}
    p(z_{t+k} | x_{1:t}, x_{t+k}) \propto p(x_{t+k} | z_{t+k}) p(z_{t+k} | x_{1:t}) \label{eq:fact}
\end{align}
using our independence assumptions in the graphical model.
Combining \eqref{eq:mi} and \eqref{eq:ce}, our final objective
for the entire utterance, is thus
\begin{align}
   \mathcal{L} &= \sum_{t=1}^{T-k} \mathbb{E}_{z_{t+k} \sim q} \big[ -\log q(z_{t+k} | x_{t+k}) \notag\\
        &\qquad {} + \log p(x_{t+k} | z_{t+k}) + \log p(z_{t+k} | x_{1:t}) \big]. \label{eq:main-loss}
\end{align}
The distributions are modeled with neural networks.
In particular, $p(z_{z_{t+k}} | x_{1:t})$ is the prediction network,
while $q(z_{t+k} | x_{t+k})$ is the confirmation network.\footnote{
The derivation here deviates slightly from
the original co-training work \cite{ref:mcallester2018information}.
We have a term that predicts the future frame, while the original work does not.
Our derivation is more aligned with variational inference.
In fact, \eqref{eq:main-loss} is exactly
the variational lower bound \cite{ref:kingma2013auto} of the log likelihood \eqref{eq:likelihood},
while $\mathcal{L} - \sum_{t=1}^{T-k} \log p(x_{t+k} | x_{1:t})$
is the lower bound of the mutual information in \eqref{eq:mi}.}

\begin{figure}[ht]
\centering\includegraphics{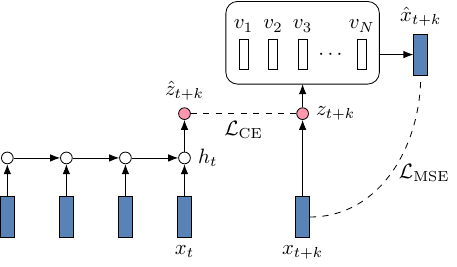}
\caption{The forward process of training.
    The recurrent neural network and the quantization layer are trained jointly
    with the co-training objective.}
\label{fig:2}
\end{figure}

For the prediction network, we choose a multilayer unidirectional recurrent network,
allowing us to compare different training objectives, such as APC and VQ-APC, while
holding the model architecture fixed.
We apply a linear projection $U$ on the output of the recurrent network to obtain
the probability of the latent variable.
Formally,
\begin{align}
    p(z_{t+k}|x_{1:t}) = p(z_{t+k} | h_t) = \frac{\exp\left(h_t^\top U \mathbf{1}_{z_{t+k}}\right)}{\sum_{j=1}^{N} \exp\left(h_t^\top u_j\right)},
\label{eq:latent}
\end{align}
where $h_t$ is the output of the recurrent network after taking $x_{1:t}$ as input,
$\mathbf{1}_{z_{t+k}}$ is a one-hot vector where the $z_{t+k}$-th element is set to 1,
and $u_j$ is the $j$-th row of $U$.

For the confirmation network, we choose
\begin{align}
    q(z_{t+k}|x_{t+k}) = \frac{\exp\left(-\|x_{t+k} - V\mathbf{1}_{z_{t+k}}\|^2\right)}
        {\sum_{j=1}^{N} \exp\left(-\|x_{t+k} - v_j\|^2\right)},
\label{eq:softmax-q}
\end{align}
where $v_j$ is the $j$-th row of a matrix $V$.
The distribution is chosen so that finding the mode of $q$
is equivalent to finding the closest row vector in $V$ to $x_{t+k}$.
Given the similarity to VQ, we call $V$ the codebook, and each row vector in $V$ a codeword.

Finally, the generation distribution of the future frame is chosen as
\begin{align}
    p(x_{t+k}|z_{t+k}) = \frac{1}{(2\pi)^{d/2}}\exp\left(-\frac{1}{2}\|x_{t+k} - W V \mathbf{1}_{z_{t+k}}\|^2\right),
\label{eq:5}
\end{align}
where $W$ is a linear projection and $d$ is the dimension of $x_{t+k}$.
In words, the probability can be interpreted as using the vector
after quantization (one of the codewords)
to generate $x_{t+k}$.
For simplicity, we let each codeword have the same dimension as
a frame and let $W$ be the identity matrix.

The parameters for training are the parameters in the recurrent network, $U$, $V$,
and $W$ (if it weren't fixed to the identity matrix).
Note that based on the choice of distribution above,
$\mathbb{E}_{z_{t+k} \sim q}[-\log p(z_{t+k} | x_{1:t})]$ is a cross-entropy (CE) loss,
$\mathbb{E}_{z_{t+k} \sim q}[-\log p(x_{t+k} | z_{t+k})]$ is a weighted mean-squared-error (MSE) loss,
and $\mathbb{E}_{z_{t+k} \sim q}[-\log q(z_{t+k} | x_{t+k})]$ is the entropy.
The forward process is shown in Figure~\ref{fig:2}.

\section{Related Work}

Based on the loss functions and Figure~\ref{fig:2}, it is now clear that
HuBERT-like training is a special case of our proposed approach.
Specifically, the two steps
\begin{gather}
    q(z_{t+k} | x_{t+k}) = \mathbbm{1}[{z_{t+k} = \argmin_{i=1, \dots, N} \|x_{t+k} - v_i\|}] \label{eq:one-hot-q}\\
    \min_V \sum_{t=1}^{T-k} \mathbb{E}_{z_{t+k} \sim q}[-\log p(x_{t+k} | z_{t+k})]
\end{gather}
where $\mathbbm{1}[s]$ is 1 if $s$ is true and 0 otherwise,
are exactly the steps in $k$-means.
If we choose \eqref{eq:one-hot-q} instead of \eqref{eq:softmax-q}
as our confirmation network and optimize the codebook $V$ first before
optimizing the recurrent network in \eqref{eq:main-loss},
then our proposed approach is exactly HuBERT-like training,
where $k$-means is first run offline and a model is
trained to predict the $k$-means centroids \cite{ref:hsu2021hubert}.\footnote{Note
the entropy of $q$ in \eqref{eq:main-loss} is 0 when $q$ is a point mass.}
This approach is only HuBERT-like, because (besides the architectural differences)
there are subsequent clustering, pseudo-labeling, and training steps
that are not covered in our approach.
Another difference is that HuBERT uses an ad-hoc loss function when
contrasting the quantized features, while our formulation simply uses
cross entropy.

Several others \cite{ref:jin2020discrete,ref:henter2018deep} have noted
VQ as a hard E-step, i.e., the choise of $q$ in \eqref{eq:one-hot-q}, in the context of variational expectation maximization,
but the two training approaches have not been empirically compared.

Our approach subsumes VQ-APC if we optimize the likelihood \eqref{eq:likelihood} directly
using Gumbel softmax to approximate the expectation \eqref{eq:future}.
Specifically, we can write \eqref{eq:future} as
\begin{align}
    p(x_{t+k} | x_{1:t}) = \mathbb{E}_{z_{t+k} \sim p(z_{t+k} | x_{1:t})} [p(x_{t+k} | z_{t+k})].
\end{align}
The expectation can be approximated by drawing a single sample from
the discrete distribution $p(z_{t+k} | x_{1:t})$.
This can be achieved by adding Gumbel noise and a temperature to \eqref{eq:latent}.
The matrix $U$ is used to compute the sampling probability,
while the matrix $V$ selects a codeword for the subsequent forward process.
The matrix $W$ is also trained, and that completes the special case of VQ-APC.

Vector quantization has been widely applied to learning discrete units
of speech \cite{ref:chorowski2019unsupervised, ref:harwath2019learning, ref:van2020vector}
There is also evidence that enforcing discreteness in the model
architecture learns a better representation for
ASR \cite{ref:baevski2020wav2vec,ref:ling2020decoar,ref:hsu2021hubert,ref:chung2021w2v}.
Each of these models is likely to have an underlying generative
process similar to the one in Figure \ref{fig:1}.
Our work makes clear the connection between the generative
process in Figure \ref{fig:1} and the forward process in Figure \ref{fig:2},
explicitly showing where the discrete variables are assumed.

We also want to emphasize that while Gumbel softmax is often
used as together with learning discrete representation,
learning discrete representation does not neccessarily need
to have Gumbel softmax.
For example, since our generation distribution $p(x_{t+k} | z_{t+k})$ is shallow,
exact marginalization is in fact efficient.
Exact marginalization becomes expensive both in time and in memory
when $p(x_{t+k} | z_{t+k})$ is deep, 
and that is when Gumbel softmax becomes neccessary, especially
when using a large number of codewords.

\section{Experiments}

\begin{figure}
    \centering
    \hspace{4.0em}
	\begin{tikzpicture}[font=\footnotesize]
	    \draw[thick, red] (0, 3.5) -- (0.3, 3.5);
        \draw[red, mark=*, mark size=1.0pt] plot coordinates {(0.15, 3.5)};
		\node[anchor=west] at (0.3, 3.5) [font=\fontsize{6.5}{0}\selectfont] {Marginalization};
	    \draw[thick, blue] (2.1, 3.5) -- (2.4, 3.5);
        \draw[blue, mark=*, mark size=1.0pt] plot coordinates {(2.25, 3.5)};
		\node[anchor=west] at (2.4, 3.5) [font=\fontsize{6.5}{0}\selectfont] {Gumbel};
	    \draw[thick, gray] (3.5, 3.5) -- (3.8, 3.5);
        \draw[gray, mark=*, mark size=1.0pt] plot coordinates {(3.65, 3.5)};
		\node[anchor=west] at (3.8, 3.5) [font=\fontsize{6.5}{0}\selectfont] {HuBERT-like};
	\end{tikzpicture}
    \includegraphics[width=23em]{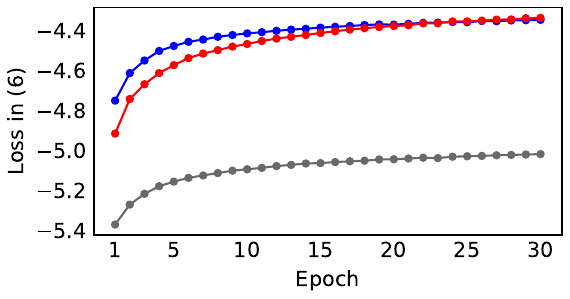}
    \caption{The co-training losses for various optimization approaches. The codebook size is $256$.
        At epoch $0$, the losses of exact marginalization, Gumbel approximation and HuBERT-like training
        are $-19.9$, $-19.9$ and $-8.4$ respectively. The epoch 0 losses are excluded for clarity.}
    \label{fig:3}
\end{figure}
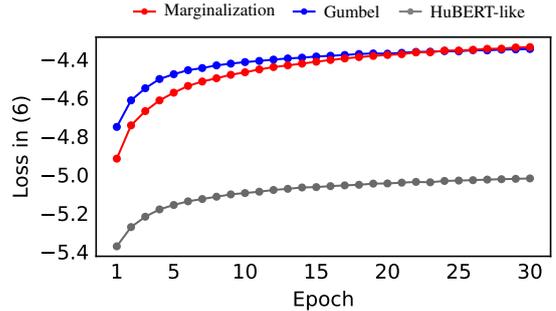

One common belief is that the learned representation through self-supervision is highly correlated
with phones \cite{ref:baevski2019vq,ref:chung2020vector,ref:van2020vector}.
To test this hypothesis, we pre-train our generative model on the 360-hour subset of LibriSpeech
and perform phone classification on Wall Street Journal (WSJ).
On top of this hypothesis, since our proposed approach subsumes
HuBERT-like training and VQ-APC as special cases, we study the effect
of different training approaches while holding the architecture fixed.

We extract 40-dimensional log Mel features for all data sets with a window size
of 25 ms and a hop length of 10 ms.
We normalize the acoustic frames based on the mean and variance of individual
training sets.
Following the protocol of prior work \cite{ref:chung2019unsupervised,ref:chung2020vector,ref:liu2020non},
we leave 10\% of the training set \texttt{si284} for development,
and report phone error rates (PER) on the test set \texttt{dev93}.
We use forced alignments obtained from a speaker-adaptive GMM-HMM as the ground truth.

We choose a 512-dimensional 3-layer unidirectional LSTM as the prediction network.
We let the time shift $k = 5$.
For the confirmation network, we only use a codebook of size $N$,
where each codeword is a 40-dimensional vector, the same dimension as the
input frames.
Regarding VQ-APC, we use a codebook of size $N$ with 512-dimensional codewords.
For the targets of HuBERT-like training, we follow prior work \cite{ref:hsu2021hubert}
and use the $k$-means algorithm with $N$ centroids.
We initialize the centroids with $k$-means++ \cite{ref:arthur2006k}
on randomly selected 3,000 utterances.
Regular $k$-means is trained for 10 iterations.
Once the models are pre-trained, we freeze the models and inspect the phonetic information in different
layers of LSTMs by adding a linear projection on top of the hidden vectors.

A learning rate of $10^{-3}$ and a batch size of 16 are used for all experiments.
We use Adam to optimize the pre-training objectives for 30 epochs,
while only using 10 epochs for the linear classifier.
Instead of using $\ell_1$ loss as in prior work \cite{ref:chung2019unsupervised,ref:chung2020vector},
our loss is MSE due to the choice of Gaussian.
For Gumbel softmax, we cool the temperature $\tau$ from $2.0$ to $0.5$ with a decay rate of $0.99995$.
Ths cooling schedule is used in both the sampling approximation of co-training and VQ-APC.
We use the straight-through estimator when computing the gradient of the sampling.

\subsection{Comparing Training Objectives}

We train three variants based on the same objective in \eqref{eq:main-loss},
HuBERT-like training, exact marginalization in \eqref{eq:main-loss}, and a Gumbel approximation
only for the term $\mathbb{E}_{z_{t+k} \sim q}[-\log p(z_{t+k} | x_{t+k})]$
to simulate the case when approximation is needed.
In HuBERT-like training, only the cross entropy term $\mathbb{E}_{z_{t+k} \sim q}[-\log p(z_{t+k} | x_{1:t})]$
is optimized, since the other two terms are already optimized with $k$-means.
We simply clamp $V$ to the $k$-means centroids when plotting the HuBERT-like
objective.
Despite using Gumbel approximation, we still report the exact marginalization
of the loss.
The training losses of the three variants are shown in Figure~\ref{fig:3}.
Since $k$-means is trained off-line for HuBERT-like, its loss starts
much higher than the other two.
However, since the other two optimizes the entire objective, their losses
rapidly surpass the HuBERT-like.
We also find that using Gumbel approximation converges faster,
but exact marginalization eventually becomes better.

\begin{table}
\begin{center}
\caption{Phone error rates (PER) on \texttt{dev93} for different learned
    representations. $N$ is the size of the codebook,
    and $h_1$, $h_2$, and $h_3$ indicate the use of the first, second,
    and third layer, respectively, for evaluation.
\label{tab:1}
}
\begin{tabular}{cc|ccc}
\toprule
\multirow{2}{*}{}                                             & \multirow{2}{*}{$N$} & \multicolumn{3}{c}{PER(\%) $\downarrow$} \\ 
                                                              &                      & $h_1$     & $h_2$             & $h_3$     \\ \midrule
APC                                                           & -                    & 30.9      & 22.1              & 24.5      \\ \midrule
\multirow{3}{*}{VQ-APC}                                       & 100                  & 28.1      & 22.4     		 & 30.0      \\
                                                              & 256                  & 27.4      & \textbf{22.0}     & 31.4      \\
                                                              & 512                  & 27.2      & 22.4              & 31.2      \\ \midrule
\multirow{3}{*}{HuBERT-like APC}                              & 100                  & 24.9      & \textbf{20.5}     & 26.2      \\
                                                              & 256                  & 23.9      & 20.6              & 27.1      \\
                                                              & 512                  & 22.6      & 21.1              & 27.9      \\ \midrule
\multirow{3}{*}{\makecell{Co-training\\(Gumbel)}}             & 100                  & 24.7      & 19.9              & 25.1      \\
                                                              & 256                  & 24.3      & \textbf{19.8}     & 25.2      \\
                                                              & 512                  & 24.9      & 19.8    			 & 25.1      \\ \midrule
\multirow{3}{*}{\makecell{Co-training\\(Marginalization)}}    & 100                  & 26.0      & 19.7              & 24.4      \\
                                                              & 256                  & 27.5      & \textbf{19.5}     & 24.2      \\
                                                              & 512                  & 26.3      & 19.5              & 24.8      \\ \bottomrule
\end{tabular}
\end{center}
\vspace{-1em}
\end{table}

\subsection{Phone Classification Results}

Given the three variants, we evaluate the learned representation with phone classification.
We also compare vanilla APC and VQ-APC given how similar the approaches are.
We use the same 3-layer unidirectional LSTM for all approaches.
We experiment three codebook sizes, namely, 100, 256, and 512.
The phone error rates (PER) are shown in Table \ref{tab:1}.

We first notice that our reimplementation of APC is better
than the prior work \cite{ref:chung2019unsupervised} due to various
differences in the training pipeline.
We observe better performance in the second layer,
consistent with the general observation that phonetic information
is more accessible in middle layers \cite{ref:chung2019unsupervised,ref:hsu2021hubert}.
HuBERT-like training is better than APC and VQ-APC,
while both the Gumbel approximation and the exact marginalization
are better than HuBERT-like.
Optimizing the co-training objective with exact marginalization
achieves an 11.8\% relative improvement over vanilla APC,
and a 4.9\% relative improvement over HuBERT-like training.

\begin{figure}[htp]
    \centering
    \includegraphics[width=24em]{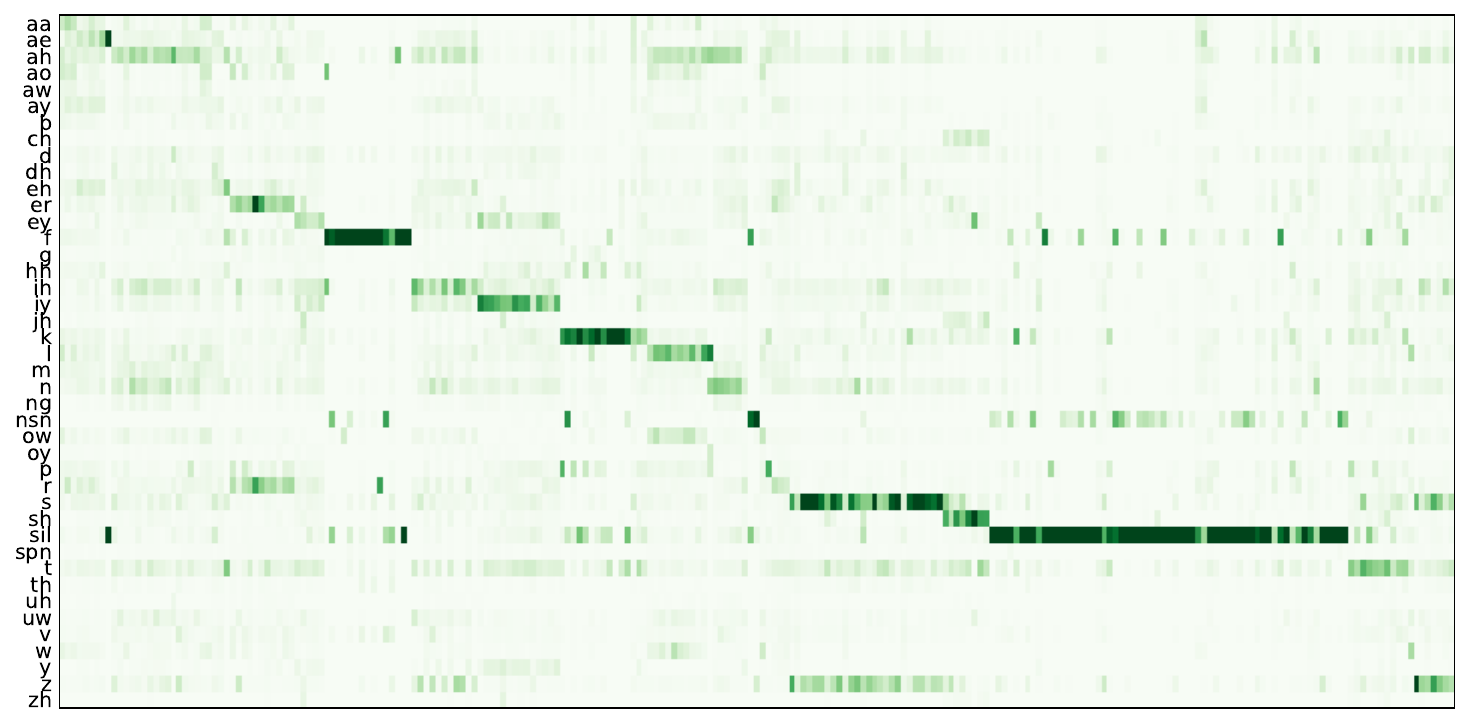}
    \includegraphics[width=24em]{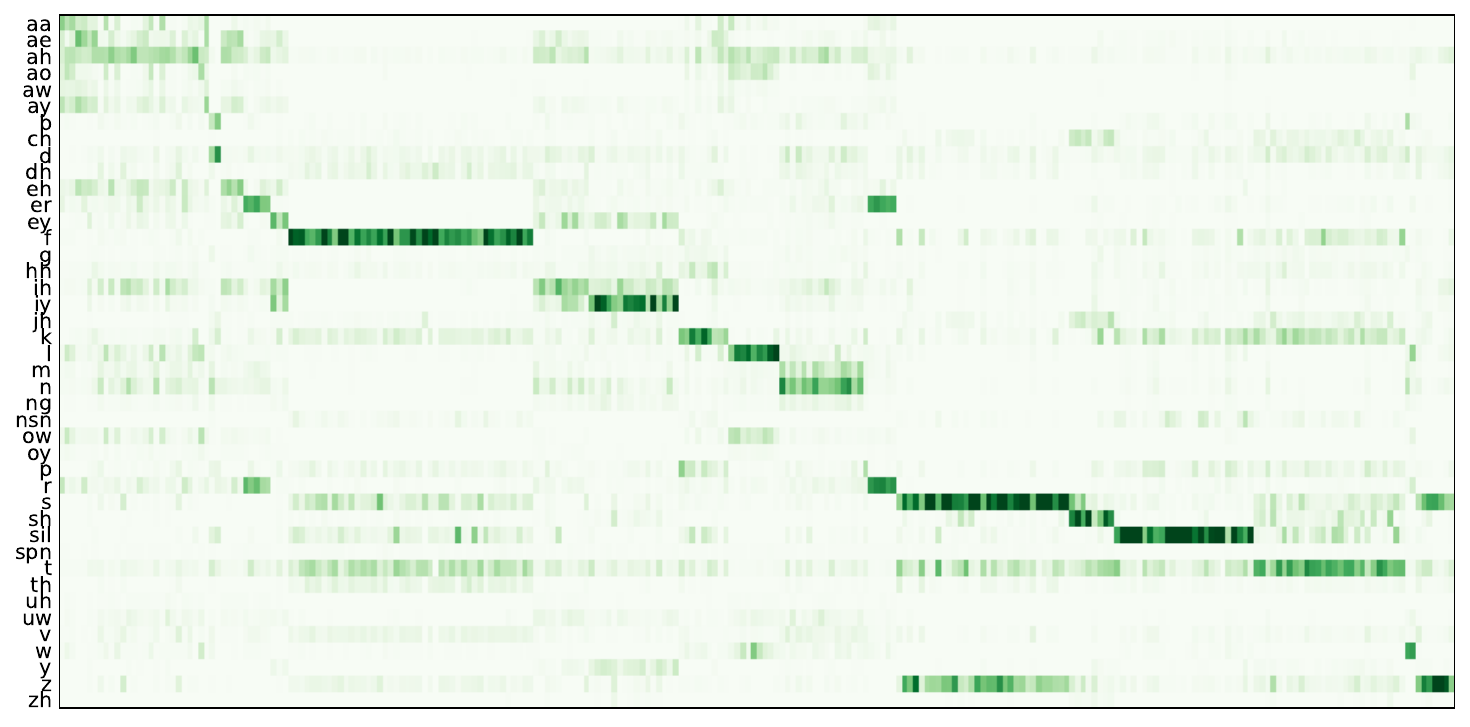}
    \caption{The conditional probabilities of phones given the codes evaluated on WSJ training set.
        The codes are based on the mode of either the prediction network (top) or confirmation network (bottom).
		The normalized mutual information is 0.228 (top) and 0.282 (bottom).
        The y-axis and x-axis represent the 42 phones and the codes respectively; the codebook size is $256$.}
    \label{fig:4}
\end{figure}

\subsection{Correlation between phones and codes}

Given the strong performance of phone classification, we expect the
discrete representation (the codes in the codebook) to correlate well with phones.
In Figure \ref{fig:4}, we plot the conditional probabilities of phones given
the codes (for $N = 256$) based on co-occurrences on the WSJ training set.
The codes are inferred based on either the mode of the
prediction network \eqref{eq:latent} or the mode of the confirmation
network \eqref{eq:softmax-q}.
We observe a strong correlation between phones and codes for both.
The codes correlate especially well with unvoiced fricatives,
namely /f, s, sh/, and silences.
The codes are also shared reasonably across phone classes,
for example, between /s/ and /z/ for similar cutoff frequency,
/r/ and /er/ for the distinct low F3,
and /iy/ and /ey/ for sharing the high F2.

\section{Conclusions}

We have shown the strong performance of autoregressive co-training
on learning discrete speech representation.
Despite the simplicity of HuBERT-like training,
predicting quantized acoustic frames alone gives a significant gain
over APC.
Optimizing the co-training objective gives the most significant gain.
Our approach subsumes HuBERT-like training and VQ-APC,
yet its generality has not been fully explored.
Viable directions for future work include using a deep confirmation
network and extending to discrete structures, similar
to \cite{ref:bhati2021segmental,ref:chorowski2021aligned}.

\bibliographystyle{IEEEtran}
\bibliography{mybib}

\begin{thebibliography}{10}
\providecommand{\url}[1]{#1}
\csname url@samestyle\endcsname
\providecommand{\newblock}{\relax}
\providecommand{\bibinfo}[2]{#2}
\providecommand{\BIBentrySTDinterwordspacing}{\spaceskip=0pt\relax}
\providecommand{\BIBentryALTinterwordstretchfactor}{4}
\providecommand{\BIBentryALTinterwordspacing}{\spaceskip=\fontdimen2\font plus
\BIBentryALTinterwordstretchfactor\fontdimen3\font minus
  \fontdimen4\font\relax}
\providecommand{\BIBforeignlanguage}[2]{{%
\expandafter\ifx\csname l@#1\endcsname\relax
\typeout{** WARNING: IEEEtran.bst: No hyphenation pattern has been}%
\typeout{** loaded for the language `#1'. Using the pattern for}%
\typeout{** the default language instead.}%
\else
\language=\csname l@#1\endcsname
\fi
#2}}
\providecommand{\BIBdecl}{\relax}
\BIBdecl

\bibitem{ref:chorowski2019unsupervised}
J.~Chorowski, R.~J. Weiss, S.~Bengio, and A.~van~den Oord, ``Unsupervised
  speech representation learning using wavenet autoencoders,'' \emph{IEEE
  Transactions on Audio, Speech, and Language Processing}, 2019.

\bibitem{ref:chung2020vector}
Y.~Chung, H.~Tang, and J.~R. Glass, ``Vector-quantized autoregressive
  predictive coding,'' in \emph{Interspeech}, 2020.

\bibitem{ref:baevski2020wav2vec}
A.~Baevski, Y.~Zhou, A.~Mohamed, and M.~Auli, ``wav2vec 2.0: {A} framework for
  self-supervised learning of speech representations,'' 2020.

\bibitem{ref:zhou2021comparison}
H.~Zhou, A.~Baevski, and M.~Auli, ``A comparison of discrete latent variable
  models for speech representation learning,'' in \emph{ICASSP}, 2021.

\bibitem{ref:hsu2021hubert}
W.~Hsu, B.~Bolte, Y.~H. Tsai, K.~Lakhotia, R.~Salakhutdinov, and A.~Mohamed,
  ``{HuBERT}: Self-supervised speech representation learning by masked
  prediction of hidden units,'' \emph{IEEE Transactions on Audio, Speech, and
  Language Processing}, 2021.

\bibitem{ref:ling2020decoar}
S.~Ling and Y.~Liu, ``{D}e{C}o{AR} 2.0: Deep contextualized acoustic
  representations with vector quantization,'' 2021.

\bibitem{ref:chung2021w2v}
Y.~Chung, Y.~Zhang, W.~Han, C.~Chiu, J.~Qin, R.~Pang, and Y.~Wu, ``w2v-{BERT}:
  Combining contrastive learning and masked language modeling for
  self-supervised speech pre-training,'' in \emph{ASRU}, 2021.

\bibitem{ref:mnih2014neural}
A.~Mnih and K.~Gregor, ``Neural variational inference and learning in belief
  networks,'' in \emph{ICML}, 2014.

\bibitem{ref:van2017neural}
A.~van~den Oord, O.~Vinyals, and K.~Kavukcuoglu, ``Neural discrete
  representation learning,'' in \emph{NeurIPS}, 2017.

\bibitem{ref:maddison2016concrete}
C.~J. Maddison, A.~Mnih, and Y.~W. Teh, ``The concrete distribution: {A}
  continuous relaxation of discrete random variables,'' in \emph{ICLR}, 2017.

\bibitem{ref:jang2016categorical}
E.~Jang, S.~Gu, and B.~Poole, ``Categorical reparameterization with
  {G}umbel-softmax,'' in \emph{ICLR}, 2017.

\bibitem{ref:mcallester2018information}
D.~McAllester, ``Information theoretic co-training,'' \emph{arXiv:1802.07572},
  2018.

\bibitem{ref:chung2019unsupervised}
Y.~Chung, W.~Hsu, H.~Tang, and J.~R. Glass, ``An unsupervised autoregressive
  model for speech representation learning,'' in \emph{Interspeech}, 2019.

\bibitem{ref:maas2017building}
A.~L.Maas, P.~Qi, Z.~Xie, A.~Y. Hannun, C.~T. Lengerich, D.~Jurafsky, and
  A.~Y.Ng, ``Building {DNN} acoustic models for large vocabulary speech
  recognition,'' \emph{Computer Speech \& Language}, 2017.

\bibitem{ref:chung2020generative}
Y.-A. Chung and J.~Glass, ``Generative pre-training for speech with
  autoregressive predictive coding,'' in \emph{ICASSP}, 2020.

\bibitem{ref:kingma2013auto}
D.~P. Kingma and M.~Welling, ``Auto-encoding variational {Bayes},'' in
  \emph{ICLR}, 2014.

\bibitem{ref:jin2020discrete}
S.~Jin, S.~Wiseman, K.~Stratos, and K.~Livescu, ``Discrete latent variable
  representations for low-resource text classification,'' in \emph{ACL}, 2020.

\bibitem{ref:henter2018deep}
G.~E. Henter, X.~Wang, and J.~Yamagishi, ``Deep encoder-decoder models for
  unsupervised learning of controllable speech synthesis,''
  \emph{arXiv:1807:11470}, 2018.

\bibitem{ref:harwath2019learning}
D.~Harwath, W.-N. Hsu, and J.~Glass, ``Learning hierarchical discrete
  linguistic units from visually-grounded speech,'' in \emph{ICLR}, 2019.

\bibitem{ref:van2020vector}
B.~van Niekerk, L.~Nortje, and H.~Kamper, ``Vector-quantized neural networks
  for acoustic unit discovery in the {ZeroSpeech 2020 Challenge},'' in
  \emph{Interspeech}, 2020.

\bibitem{ref:baevski2019vq}
A.~Baevski, S.~Schneider, and M.~Auli, ``vq-wav2vec: Self-supervised learning
  of discrete speech representations,'' in \emph{ICLR}, 2020.

\bibitem{ref:liu2020non}
A.~H. Liu, Y.~Chung, and J.~R. Glass, ``Non-autoregressive predictive coding
  for learning speech representations from local dependencies,'' in
  \emph{Interspeech}, 2021.

\bibitem{ref:arthur2006k}
D.~Arthur and S.~Vassilvitskii, ``k-means++: the advantages of careful
  seeding,'' in \emph{SODA}, 2007.

\bibitem{ref:bhati2021segmental}
S.~Bhati, J.~Villalba, P.~Zelasko, L.~Moro{-}Vel{\'{a}}zquez, and N.~Dehak,
  ``Segmental contrastive predictive coding for unsupervised word
  segmentation,'' in \emph{Interspeech}, 2021.

\bibitem{ref:chorowski2021aligned}
J.~Chorowski, G.~Ciesielski, J.~Dzikowski, A.~Lancucki, R.~Marxer, M.~Opala,
  P.~Pusz, P.~Rychlikowski, and M.~Stypulkowski, ``Aligned contrastive
  predictive coding,'' in \emph{Interspeech}, 2021.

\end{thebibliography}

\end{document}